\documentclass{article}

\usepackage{graphicx}
\usepackage{fancyhdr}
\usepackage[final]{self-hpo}
\usepackage{microtype}
\usepackage{times}
\usepackage[T1]{fontenc}
\usepackage{lineno}
\usepackage{fontawesome5}
\usepackage{marvosym}

\usepackage[most]{tcolorbox}
\definecolor{SoftBlue}{RGB}{232,242,255}
\newtcolorbox{abstractbox}{
  colback=SoftBlue,
  colframe=SoftBlue,
  boxrule=0pt,
  arc=2mm,
  left=4mm,right=4mm,top=2mm,bottom=2mm
}

\usepackage{amsmath,amsfonts,bm}

\def\eqref#1{equation~\ref{#1}}

\def\1{\bm{1}}

\DeclareMathAlphabet{\mathsfit}{\encodingdefault}{\sfdefault}{m}{sl}
\SetMathAlphabet{\mathsfit}{bold}{\encodingdefault}{\sfdefault}{bx}{n}

\usepackage{hyperref}
\usepackage{url}
\usepackage{graphicx}
\usepackage{booktabs}
\usepackage{amssymb}
\usepackage{pifont}
\usepackage{multirow}
\usepackage{subcaption}
\usepackage{xcolor}
\usepackage{colortbl}
\usepackage{xspace}
\usepackage{float}
\usepackage{placeins}
\usepackage{wrapfig}
\usepackage{algorithm}
\usepackage{algpseudocode}
\usepackage{tikz}
\usetikzlibrary{arrows.meta,positioning,fit,calc}
\usepackage{pgfplots}
\pgfplotsset{compat=1.18}

\hypersetup{
    colorlinks,
    linkcolor={red!50!black},
    citecolor={blue!50!black},
    urlcolor={blue!80!black}
}

\usepackage[most]{tcolorbox}
\definecolor{hOld}{HTML}{CF222E}\definecolor{hOldBg}{HTML}{FFF5F5}\definecolor{hOldTitle}{HTML}{FDE2E2}
\definecolor{hNew}{HTML}{1A7F37}\definecolor{hNewBg}{HTML}{F4FBF6}\definecolor{hNewTitle}{HTML}{DDF2E3}
\definecolor{hKw}{HTML}{1F2328}\definecolor{hShared}{HTML}{0969DA}\definecolor{cPTA}{HTML}{0969DA}\definecolor{cStep}{HTML}{BC4C00}\definecolor{cExec}{HTML}{1B7C83}\definecolor{cQuery}{HTML}{8250DF}\definecolor{cCheck}{HTML}{BF3989}\definecolor{cSub}{HTML}{9A6700}\definecolor{hStr}{HTML}{0A3069}\definecolor{hCm}{HTML}{6E7781}
\newcommand{\cfn}[2]{\textcolor{#1}{\fontencoding{T1}\fontfamily{lmtt}\selectfont\fontsize{5.6}{6.8}\selectfont #2}}
\newcommand{\bad}[1]{\textcolor{hOld}{\normalfont\sffamily\fontsize{5.6}{6.8}\selectfont\ding{55}~#1}}
\newcommand{\good}[1]{\textcolor{hNew}{\normalfont\sffamily\fontsize{5.6}{6.8}\selectfont\ding{51}~#1}}
\tcbset{codebox/.style={listing only, enhanced, boxrule=0.5pt, arc=2pt, left=4pt, right=3pt, top=2pt, bottom=2pt,
  toptitle=1.5pt, bottomtitle=1.5pt, coltitle=black, fonttitle=\footnotesize\bfseries,
  listing options={language=Python, basicstyle=\fontencoding{T1}\fontfamily{lmtt}\selectfont\fontsize{5.6}{6.8}\selectfont, keywordstyle=\color{hKw}\bfseries,
    stringstyle=\color{hStr}, commentstyle=\color{hCm}, columns=fullflexible, keepspaces=true,
    showstringspaces=false, emph={[1]ProgressTrackingAgent}, emphstyle={[1]\color{cPTA}}, emph={[2]step}, emphstyle={[2]\color{cStep}}, emph={[3]execute_actions}, emphstyle={[3]\color{cExec}}, emph={[4]query}, emphstyle={[4]\color{cQuery}}, emph={[5]_check_finished}, emphstyle={[5]\color{cCheck}}, emph={[6]Submitted}, emphstyle={[6]\color{cSub}}, escapechar=|}}}
\tcbset{fullcode/.style={listing only, enhanced, breakable, boxrule=0.5pt, arc=2pt, left=4pt, right=3pt, top=2pt, bottom=2pt,
  toptitle=1.5pt, bottomtitle=1.5pt, coltitle=black, fonttitle=\footnotesize\bfseries, colframe=hNew!60, colback=hNewBg, colbacktitle=hNewTitle,
  listing options={language=Python, basicstyle=\fontencoding{T1}\fontfamily{lmtt}\selectfont\fontsize{5.8}{7}\selectfont, keywordstyle=\color{hKw}\bfseries,
    stringstyle=\color{hStr}, commentstyle=\color{hCm}, columns=fullflexible, keepspaces=true, breaklines=true,
    showstringspaces=false, emph={[1]ProgressTrackingAgent}, emphstyle={[1]\color{cPTA}}, emph={[2]step}, emphstyle={[2]\color{cStep}}, emph={[6]Submitted}, emphstyle={[6]\color{cSub}}}}}
\lstdefinelanguage{pydiff}{morecomment=[l]{\#},
  moredelim=**[l][\color{hNew}]{+},moredelim=**[l][\color{hOld}]{-},moredelim=**[l][\color{hCm}]{@@}}
\tcbset{diffcode/.style={listing only, enhanced, breakable, boxrule=0.5pt, arc=2pt, left=4pt, right=3pt, top=2pt, bottom=2pt,
  toptitle=1.5pt, bottomtitle=1.5pt, coltitle=black, fonttitle=\footnotesize\bfseries, colframe=black!35, colback=black!2, colbacktitle=black!8,
  listing options={language=pydiff, basicstyle=\fontencoding{T1}\fontfamily{lmtt}\selectfont\fontsize{6}{7.4}\selectfont,
    columns=fullflexible, keepspaces=true, breaklines=true, showstringspaces=false, commentstyle=\color{hCm}}}}
\usepackage{fvextra}
\usepackage{needspace}
\definecolor{rUser}{HTML}{6B7280}\definecolor{rModel}{HTML}{6585BC}\definecolor{rCall}{HTML}{C8792E}
\definecolor{rResp}{HTML}{8A8F98}\definecolor{rHarness}{HTML}{1A7F37}\definecolor{rSub}{HTML}{374151}
\newcommand{\trajbox}[3]{\newtcolorbox{#1}[1][]{enhanced,breakable,sharp corners,frame hidden,boxrule=0pt,
  borderline west={1.8pt}{0pt}{#2},colback=#2!5,left=6pt,right=4pt,top=0pt,bottom=2pt,
  before skip=3pt,after skip=1pt,fontupper=\scriptsize,title={\sffamily\bfseries\fontsize{5.5}{6}\selectfont #3},
  coltitle=#2!85!black,colbacktitle=#2!5,toptitle=2pt,bottomtitle=0pt,titlerule=0pt,##1}}
\definecolor{trajNote}{HTML}{6A3FA0}
\newcommand{\trajnote}[1]{\par\vspace{2pt}\noindent{\color{trajNote}\sffamily\bfseries\footnotesize $\blacktriangleright$~#1}\par\vspace{1pt}}
\trajbox{trajusr}{rUser}{Task}
\trajbox{trajasst}{rModel}{Agent}
\trajbox{trajcall}{rCall}{Action}
\trajbox{trajresp}{rResp}{Observation}
\trajbox{trajharness}{rHarness}{Harness}
\trajbox{trajsub}{rSub}{Final submission}
\newtcolorbox[list inside=prompt,auto counter]{prompt}[1][]{
    colbacktitle=black!60,
    coltitle=white,
    fontupper=\footnotesize,
    boxsep=5pt,
    left=0pt,
    right=0pt,
    top=0pt,
    bottom=0pt,
    boxrule=1pt,
    #1,
}

\newcommand{\eg}{{\sl e.g.}}
\newcommand{\model}{\mbox{Self-HPO}\xspace}
\newcommand{\hstep}{H-step\xspace}
\newcommand{\pstep}{P-step\xspace}
\newcommand{\cmark}{\ding{51}}
\newcommand{\xmark}{\ding{56}}

\newcommand{\tabmeanstd}[2]{#1{\scriptstyle\,\pm\,#2}}
\newcommand{\tabbestmeanstd}[2]{\mathbf{#1}{\scriptstyle\,\boldsymbol{\pm}\,\mathbf{#2}}}

\title{Self-Evolving Coding Agents via Harness-Policy Optimization}

\author{%
  \textbf{Zhepei Wei}\textsuperscript{1,2}\thanks{Work done during internship at Microsoft Research.},
  \textbf{Wenlin Yao}\textsuperscript{1},
  \textbf{Baolin Peng}\textsuperscript{1},
  \textbf{Danqing Wang}\textsuperscript{3},
  \textbf{Qianhui Wu}\textsuperscript{1} \\
  \textbf{Hao Cheng}\textsuperscript{1},
  \textbf{Yu Meng}\textsuperscript{2},
  \textbf{Jianfeng Gao}\textsuperscript{1} \\
  \textsuperscript{1}Microsoft Research \quad
  \textsuperscript{2}University of Virginia \quad
  \textsuperscript{3}Carnegie Mellon University \\[2mm]
  \faGithub\
  \href{https://github.com/MSR-Orchard/Self-HPO}{https://github.com/MSR-Orchard/Self-HPO}
  \qquad
  \includegraphics[height=0.9em]{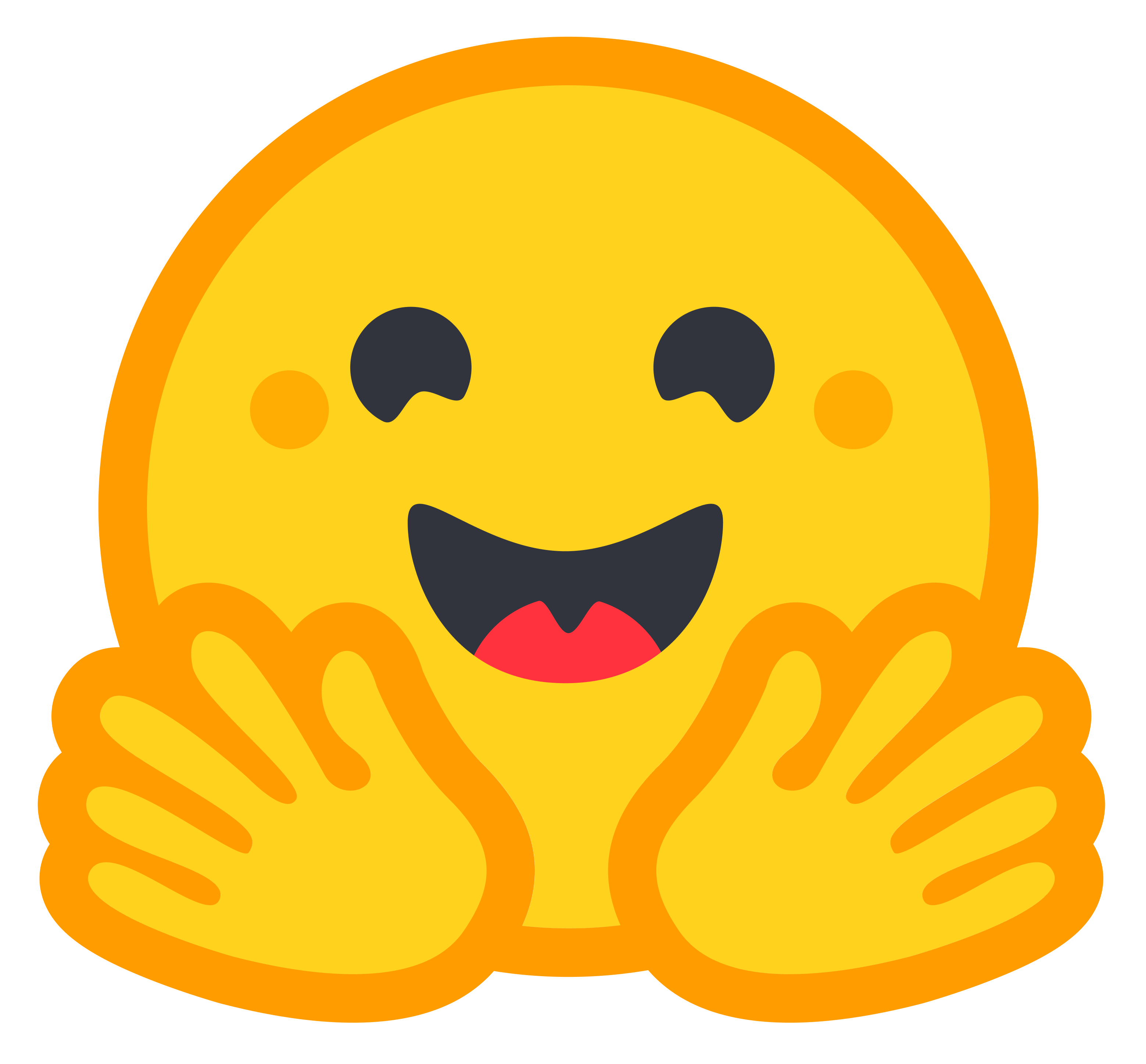}\
  \href{https://huggingface.co/self-hpo}{https://huggingface.co/self-hpo}
}

\newcommand{\dlt}[1]{\,{\scriptsize\color{red!70!black}(#1)}}

\begin{document}

\renewcommand{\thefootnote}{\arabic{footnote}}

\ifcolmsubmission
\linenumbers
\fi

\maketitle
\vspace{-1.0cm}

\begin{abstractbox}
\begin{abstract}
Large language model (LLM) agents have shown promising potential in coding tasks, and their performance is jointly determined by the policy model and the harness that mediates their interaction with the environment.
Indeed, harness and policy are mutually reinforcing and evolving only one component is suboptimal---a harness tuned for one checkpoint may not target the failure modes of an updated policy, while policy learning remains limited to the trajectories elicited by a fixed harness.
In light of this, we propose \model, a recursive self-improving framework for coding agents that alternates harness and policy optimization, using the evolving policy as its own harness proposer.
Specifically, in each round, the current policy first analyzes its own failures and proposes new harness candidates (\hstep).
The policy is then updated through reinforcement learning (RL) under the selected harness (\pstep), and hard tasks identified during training form the next harness search set, keeping harness optimization targeted to the policy's evolving failure modes.
Across three coding benchmarks, \model consistently outperforms harness-only and policy-only baselines, improving Qwen3.5-35B-A3B's average resolve rate from 57.8\% to 64.8\%, with gains generalizing to out-of-distribution settings.
Further analysis shows that both optimized components retain gains even when paired with their base counterparts, while their joint gain exceeds the sum of their individual gains.
\model also improves reliability (pass@1) and expands the capability frontier (pass@$k$) over the base agent.
Additionally, we show that with a stronger external model (\eg, Opus 4.8) as the harness proposer, the same framework yields a further performance boost.
\end{abstract}

\end{abstractbox}

\section{Introduction}
Large language model (LLM) agents have made rapid progress on complex, multi-step tasks such as software engineering~\citep{jimenez2024swebench,yang2024sweagent}, web navigation~\citep{zhou2024webarena,he2024webvoyager,wei2025webagentr1}, and computer use~\citep{xie2024osworld,agashe2025agents2}.
Such an agent consists of two components: the policy model and its harness, which is the external scaffold that decides what the model observes, how its actions are executed, what feedback it receives, and when an episode terminates~\citep{yao2023react,shinn2023reflexion,yang2024sweagent,he2024webvoyager,lee2026metaharness,zhang2026selfharness}.
Typically, the policy model defines the agent's capability boundary, while the harness determines how these capabilities are elicited, hence the same model may succeed or fail on one problem depending on its harness~\citep{le2026multiharness,zhang2026selfharness}.

There have been two main lines of research on improving such agents.
Harness optimization holds the policy fixed and searches over prompts, workflows, or agent code~\citep{hu2024adas,lee2026metaharness,zhang2026selfharness}, whereas policy optimization holds the harness fixed and updates the model from successful trajectories or verifiable rewards~\citep{shao2024deepseekmath,guo2025deepseek,wei2025swerl}.
However, harness-only optimization cannot improve the underlying model, while policy-only training can learn only from the trajectories that its fixed harness elicits.
Although recent works begin to optimize both components~\citep{kim2026whale,chen2026harnessforge}, they rely on an external harness proposer and face the challenge of a moving optimization target as the policy and its failure modes evolve, leaving self-driven evolution for coding agents underexplored.

\begin{figure*}[t]
\centering
\begin{minipage}[t]{0.68\textwidth}
\vspace{0pt}
\centering
\includegraphics[width=\linewidth]{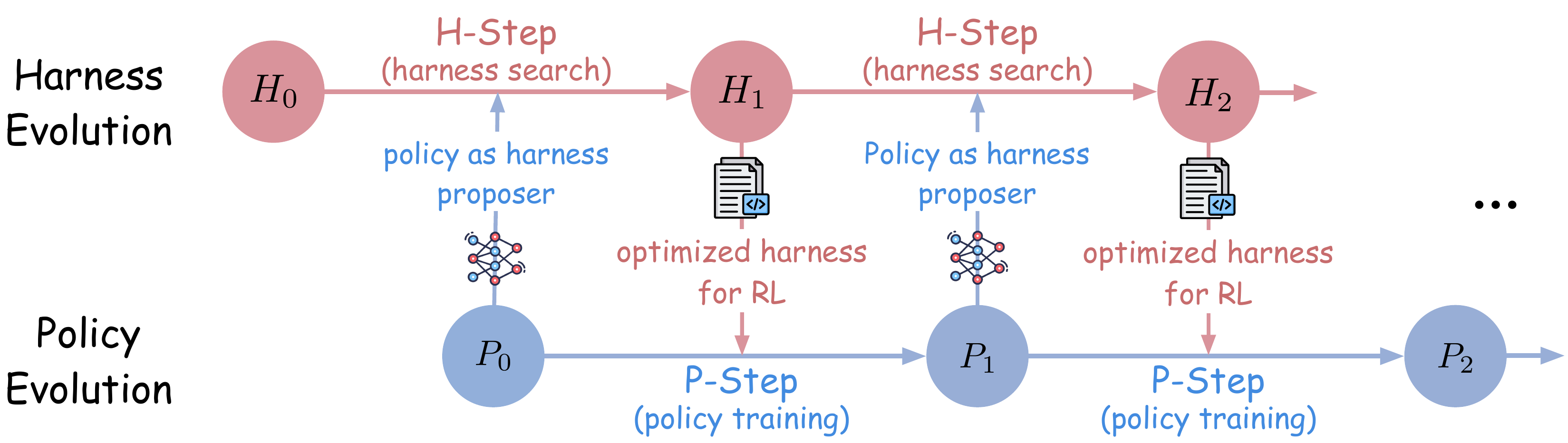}
\end{minipage}\hfill
\begin{minipage}[t]{0.30\textwidth}
\vspace{0pt}
\centering
\includegraphics[width=\linewidth]{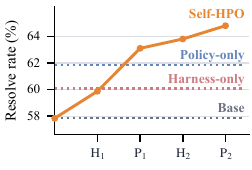}
\end{minipage}
\caption{\textbf{\model recursively improves the harness and policy using the agent itself.}
\textbf{Left:} In each \hstep, the current policy proposes an improved harness, and each \pstep then updates the policy through reinforcement learning with the optimized harness.
\textbf{Right:} Average resolve rate over SWE-Bench Verified, SWE-Bench Multilingual, and SWE-Bench Pro along the H- and P-steps. Policy-only and Harness-only denote the RLVR~\citep{peng2026orchard} and Self-Harness~\citep{zhang2026selfharness} baselines.}
\label{fig:teaser}
\vspace{-1em}
\end{figure*}

In this work, we introduce \model, a framework for recursive self-improvement driven by the agent's own policy.
As illustrated in Figure~\ref{fig:teaser}, \model alternates harness optimization (\hstep) with policy optimization under the optimized harness (\pstep).
Harness optimization exposes and compensates for the weaknesses of the current policy, while policy optimization internalizes the elicited behaviors into the model and shifts the failures that the next harness should target.
Specifically, in the \hstep, the current policy analyzes its failed trajectories and proposes harness revisions targeting its own failure modes, and selects the updated harness based on search-set performance (\S\ref{sec:hstep}).
In the \pstep, the policy is updated using reinforcement learning under the optimized harness (\S\ref{sec:pstep}).
In the next \hstep, the updated policy serves as the harness proposer, with hard tasks identified from the preceding \pstep as the search set to keep harness search aligned with the policy’s evolving failure modes.

We instantiate \model with Qwen3.5-35B-A3B~\citep{qwen2026qwen35} as the base policy, and mini-SWE-agent~\citep{yang2024sweagent} as the seed harness.
Across three coding benchmarks, \model consistently outperforms policy-only and harness-only optimization, with the average resolve rate improving at every optimization step, as demonstrated in Figure~\ref{fig:teaser}.
On SWE-Bench Verified~\citep{chowdhury2024swebenchverified}, it improves the base resolve rate from 61.0\% to 69.5\%, with transferable gains on the out-of-distribution SWE-Bench Multilingual~\citep{yang2025swesmith} and SWE-Bench Pro~\citep{deng2025swebenchpro}.
Ablations further show that evolving the harness search set with hard tasks identified during policy learning yields better performance than reusing a fixed set.
We also extend \model to the external-model setup where Opus 4.8 is used as an external harness proposer, which further raises the resolve rate on SWE-Bench Verified to 71.6\%.
The main contributions of this work are as follows:
\begin{itemize}
    \item We propose \model, a recursive self-improving framework for coding agents that alternates harness and policy optimization, where the agent uses its own policy as the harness proposer and adapts harness search to the policy's evolving failure modes.
    \item We demonstrate that \model consistently outperforms policy-only and harness-only baselines across three coding benchmarks, improving the average resolve rate from 57.8\% to 64.8\%, with gains generalizing to out-of-distribution settings.
    \item Our study shows that both optimized components of \model retain gains even when paired with their base counterparts, while their joint gain exceeds the sum of their individual gains. We also find that \model not only improves reliability (pass@1) but also expands the capability frontier (pass@$k$) of the agent.
\end{itemize}

\section{Self-HPO: Self-Evolving via Harness-Policy Optimization}
\label{sec:method}

\begin{figure*}[t]
\centering
\includegraphics[width=\textwidth]{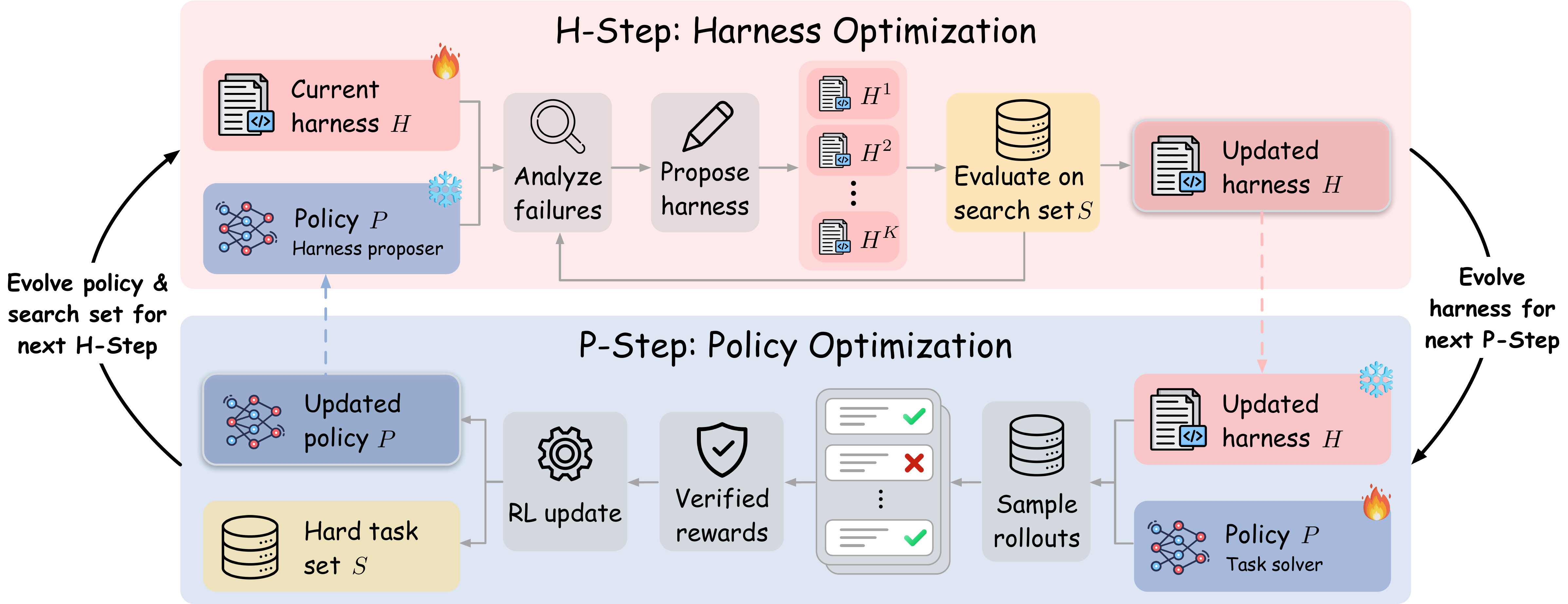}
\caption{\textbf{Overview of Self-HPO.}
\textbf{Top (\hstep):} The frozen policy analyzes the agent's failures on the search set and proposes new harness candidates.
\textbf{Bottom (\pstep):} With the updated harness, the policy evolves through reinforcement learning.
The updated policy becomes the next harness proposer, while tasks never resolved during this \pstep form the next search set.
}
\label{fig:method_overview}
\vspace{-0.5em}
\end{figure*}

\subsection{Problem Setup}
\label{sec:problem_setup}

Given a software-engineering task $x$, a coding agent combines a policy $P$, parameterized by an LLM, with a harness $H$, which controls context, tools, feedback, and termination.
The resulting trajectory $\tau\sim\pi_{H,P}(\cdot\mid x)$ is scored by an executable verifier $y(x,\tau)\in\{0,1\}$.
We measure a harness--policy pair by its expected resolve rate on the training data $\mathcal{D}$:
\begin{equation}
J(H,P)=\mathbb{E}_{x\sim\mathcal{D}}\,\mathbb{E}_{\tau\sim\pi_{H,P}(\cdot\mid x)}\big[y(x,\tau)\big].
\label{eq:objective}
\end{equation}

We study the problem of self-improving a coding agent's own harness and policy. The harness changes which behaviors the current policy can reliably express, while a policy update changes the failures that the next harness should target.
As shown in Figure~\ref{fig:method_overview}, each \hstep holds the policy fixed while using it to analyze failures and propose harness revisions, and
each \pstep then freezes this harness and updates the policy through reinforcement learning.
The updated policy becomes the next harness proposer, with hard tasks identified from the training process forming the next search set.

We call this alternating framework Harness-Policy Optimization (HPO) and implement it in two setups.
In the self-improving setup (\model), the evolving policy proposes its own harness candidates, so no separate proposer model is needed.
For reference, we also instantiate the same HPO loop with Opus 4.8 as a fixed external proposer.

\FloatBarrier
\begin{algorithm}[t]
\caption{Self Harness-Policy Optimization (Self-HPO)}
\label{alg:harmo}
\small
\algtext*{EndFor}
\algtext*{EndWhile}
\algtext*{EndIf}
\begin{algorithmic}[1]
\Require Seed harness $H_0$, base policy $P_0$, training data $\mathcal{D}$, harness search budget $K$
\State $t\gets 0$, randomly initialize harness search set $\mathcal{S}_t\subset\mathcal{D}$ 
\While{not converged} \Comment{\textcolor{blue}{Stopping criterion (\S\ref{sec:method})}}
    \Statex \textcolor{blue}{\texttt{/* H-step: search harness with the current policy as proposer */}}
    \State Score $v_0$, trajectories $\mathcal{T}_0\gets\operatorname{Eval}(H_t,P_t,\mathcal{S}_{t})$, experience $\mathcal{E}\gets\{(H_t,\mathcal{T}_0)\}$, $v_{\mathrm{best}}\gets v_0$, $H_{t+1}\gets H_t$ 

    \For{$k=1,\ldots,K$}
        \State $H_t^k\gets\operatorname{ProposeHarness}(H_{t+1},P_t,\mathcal{E})$ \Comment{\textcolor{blue}{Self-proposer (\S\ref{sec:hstep})}}
        \State $v_k, \mathcal{T}_{k}\gets\operatorname{Eval}(H_t^k,P_t,\mathcal{S}_{t})$, $\mathcal{E}\gets\mathcal{E}\cup\{(H_t^k,\mathcal{T}_k)\}$
        \If{$v_k>v_{\mathrm{best}}$}
            \State $H_{t+1}\gets H_t^k$, $v_{\mathrm{best}}\gets v_k$
        \EndIf
    \EndFor
    \Statex \textcolor{blue}{\texttt{/* P-step: train the policy under the updated harness */}}
    \State $P_{t+1},\mathcal{T}_{\mathrm{RL}}\gets\operatorname{RL}(H_{t+1},P_t,\mathcal{D})$
    \State $\mathcal{S}_{t+1}\gets\operatorname{SelectHardTasks}(\mathcal{T}_{\mathrm{RL}})$ \Comment{\textcolor{blue}{Evolving search set (\S\ref{sec:pstep})}}
    \State $t\gets t+1$

\EndWhile
\State \Return $H_t$, $P_t$
\end{algorithmic}
\end{algorithm}

A formal description of \model is presented in Algorithm~\ref{alg:harmo}.
Starting from a base policy $P_0$, a seed harness $H_0$, and a random search set $\mathcal{S}_0\subset\mathcal{D}$, the method alternates two updates.
At round $t$ (starting from $t=0$), the \hstep keeps $P_t$ fixed and uses it to propose $K$ harness candidates $\{H_t^k\}_{k=1}^{K}$, retaining the best-scoring harness on $\mathcal{S}_t$ as $H_{t+1}$.
The \pstep then freezes $H_{t+1}$ and updates $P_t$ to $P_{t+1}$ on $\mathcal{D}$ using reinforcement learning.
The resulting training rollouts identify the next hard-task set $\mathcal{S}_{t+1}$ and the loop stops when the hard-task set stabilizes between rounds.

\subsection{\hstep: Failure-Conditioned Harness Search}
\label{sec:hstep}
\paragraph{Self-proposer for harness optimization.}
The \hstep uses the frozen $P_t$ as both task solver and harness proposer.
The current pair $(H_t,P_t)$ first generates trajectories on $\mathcal{S}_t$, and the proposer inspects the current failures and produces targeted revisions to the prompt or executable runtime logic (see prompt in Appendix~\ref{app:prompts}).
Each of the $K$ search iterations revises the current best harness, starting from $H_t$, using the accumulated experience of all evaluated harnesses and their trajectories.
All candidates are evaluated on the same search set $\mathcal{S}_t$.

\paragraph{Evolving search set.}
The search set remains fixed within an \hstep but evolves between rounds.
For initialization, we sample $\mathcal{S}_0$ uniformly from $\mathcal{D}$. Subsequent search sets consist of hard tasks identified from the training trajectories generated during the preceding \pstep, where a task is considered hard if it is never resolved in any of its rollouts in that \pstep.
The next \hstep regenerates trajectories on those hard tasks with the updated harness--policy pair, producing fresh failure evidence for proposal.
Harness search therefore follows the failures exposed by the preceding policy update instead of repeatedly optimizing against the initial tasks.

\subsection{\pstep: Policy Optimization under the Optimized Harness}
\label{sec:pstep}
The \pstep improves the policy through reinforcement learning under the optimized harness $H_{t+1}$.
Starting from $P_t$, we collect on-policy trajectories over $\mathcal{D}$ as the policy is updated.
An executable verifier assigns binary outcome rewards, and our default optimizer combines Group Relative Policy Optimization (GRPO)~\citep{shao2024deepseekmath} with Balanced Adaptive Rollout (BAR) from Orchard~\citep{peng2026orchard}.
BAR adaptively draws additional trajectories to construct reward-balanced groups, reducing uninformative zero-variance groups under sparse outcome rewards, and GRPO then updates the policy using the resulting group-relative advantages.
The tasks that were never resolved during this \pstep are identified as the hard-task set $\mathcal{S}_{t+1}$, which will be passed together with the updated policy $P_{t+1}$ to the next \hstep.

\section{Experiments}
\label{sec:experiments}

\subsection{Experimental Setup}
\label{sec:setup}

\paragraph{Harness, model, and data.}
All experiments use Qwen3.5-35B-A3B~\citep{qwen2026qwen35} as the base policy and mini-SWE-agent~\citep{yao2023react,yang2024sweagent} as the seed harness.
Following Orchard~\citep{peng2026orchard}, we construct a training set of 2{,}484 tasks from SWE-rebench~\citep{badertdinov2025swerebench}, SWE-rebench V2~\citep{badertdinov2026swerebench}, and Scale-SWE~\citep{zhao2026immersion}, which is used for both harness search and policy optimization. 
We perform two rounds of alternating harness and policy optimization based on the stopping criterion, with $K=6$ harness candidates per H-step and 50 RL updates per P-step.
Appendix~\ref{app:reproducibility} describes the implementation details.

\paragraph{Benchmarks and evaluation metrics.}
SWE-Bench Verified (SWE-Verified) contains 500 human-validated GitHub issues and is our primary benchmark~\citep{jimenez2024swebench,chowdhury2024swebenchverified}.
We further evaluate out-of-distribution performance on SWE-Bench Multilingual (SWE-Multi; 300 tasks)~\citep{yang2025swesmith} and the public V2 set of SWE-Bench Pro (SWE-Pro; 642 tasks)~\citep{deng2025swebenchpro}, covering other programming languages and more diverse repository-level tasks.
We report resolve rate, the percentage of tasks whose generated patches pass the benchmark verifier.
Main results are averaged over three evaluation runs (avg@3) with standard deviations.

\paragraph{Baselines.}
We compare our method with policy-only and harness-only baselines under two setups, with the base agent as a reference.
In the \emph{self-improving setup}, RLVR~\citep{peng2026orchard} updates the policy under the seed harness, whereas Self-Harness~\citep{zhang2026selfharness} optimizes the harness using the base policy as proposer without updating its weights.
Self-HPO alternates both updates and uses the current policy to propose harness changes.
In the \emph{external-model setup}, supervised fine-tuning (SFT) learns from resolved Opus 4.8 trajectories on the same training tasks, while Meta-Harness~\citep{lee2026metaharness} uses Opus 4.8 to optimize the harness with the target policy fixed.
HPO combines external harness proposals with RL policy updates.

\begin{table}[t!]
\centering
\caption{\textbf{Main results on in-distribution (SWE-Bench Verified) and out-of-distribution evaluations (SWE-Bench Multilingual and SWE-Bench Pro).}
The optimization columns indicate whether each method updates the harness or policy.
We report the mean resolve rates with standard deviations over three runs. Bold marks the best result within each setup.}
\label{tab:main_comparison}
\small
\setlength{\tabcolsep}{2.8pt}
\renewcommand{\arraystretch}{1.08}
\resizebox{\textwidth}{!}{%
\begin{tabular}{lccccc>{\columncolor{gray!15}}c}
\toprule
\textbf{Method} & \textbf{Harness Opt} & \textbf{Policy Opt}
    & \textbf{SWE-Verified}
    & \textbf{SWE-Multi}
    & \textbf{SWE-Pro}
    & \textbf{Avg.} \\
\midrule
Base
    & {\color{red!60!black}\xmark} & {\color{red!60!black}\xmark}
    & $\tabmeanstd{61.0}{0.2}$
    & $\tabmeanstd{54.2}{0.3}$
    & $\tabmeanstd{58.3}{1.2}$
    & $\tabmeanstd{57.8}{0.4}$ \\
\midrule
\multicolumn{7}{c}{\textit{Setting 1: External-model setup (Opus 4.8)}}\\
\shortstack[l]{SFT}
    & {\color{red!60!black}\xmark} & {\color{green!60!black}\cmark}
    & $\tabmeanstd{65.0}{1.1}$
    & $\tabmeanstd{49.6}{0.8}$
    & $\tabmeanstd{49.1}{0.6}$
    & $\tabmeanstd{54.6}{0.5}$ \\
Meta-Harness~\citep{lee2026metaharness}
    & {\color{green!60!black}\cmark} & {\color{red!60!black}\xmark}
    & $\tabmeanstd{69.1}{1.9}$
    & $\tabmeanstd{56.5}{0.9}$
    & $\tabmeanstd{64.8}{1.4}$
    & $\tabmeanstd{63.5}{0.8}$ \\
{\bf HPO (ours)}
    & {\color{green!60!black}\cmark} & {\color{green!60!black}\cmark}
    & $\tabbestmeanstd{71.6}{0.9}$
    & $\tabbestmeanstd{61.4}{0.7}$
    & $\tabbestmeanstd{66.0}{0.5}$
    & $\tabbestmeanstd{66.3}{0.4}$ \\
\midrule
\multicolumn{7}{c}{\textit{Setting 2: Self-improving setup}}\\
RLVR~\citep{peng2026orchard}
    & {\color{red!60!black}\xmark} & {\color{green!60!black}\cmark}
    & $\tabmeanstd{67.5}{0.8}$
    & $\tabmeanstd{56.3}{1.0}$
    & $\tabmeanstd{61.8}{1.3}$
    & $\tabmeanstd{61.9}{0.6}$ \\
Self-Harness~\citep{zhang2026selfharness}
    & {\color{green!60!black}\cmark} & {\color{red!60!black}\xmark}
    & $\tabmeanstd{65.2}{1.5}$
    & $\tabmeanstd{54.3}{0.7}$
    & $\tabmeanstd{60.7}{0.9}$
    & $\tabmeanstd{60.1}{0.6}$ \\
\textbf{Self-HPO (ours)}
    & {\color{green!60!black}\cmark} & {\color{green!60!black}\cmark}
    & $\tabbestmeanstd{69.5}{0.9}$ 
    & $\tabbestmeanstd{59.7}{0.8}$
    & $\tabbestmeanstd{65.2}{1.4}$
    & $\tabbestmeanstd{64.8}{0.6}$ \\
\bottomrule
\end{tabular}}
\end{table}

\subsection{Main Results}
\label{sec:main_results}

\paragraph{\model consistently outperforms policy-only and harness-only optimization.}
As shown in Table~\ref{tab:main_comparison}, jointly optimizing the harness and policy achieves the best performance on all three benchmarks within each setup.
Without an external proposer, Self-HPO reaches $69.5\%$ on SWE-Verified, outperforming RLVR and Self-Harness by $2.0$ and $4.3$ points, respectively.
Its average score improves from $57.8\%$ for Base to $64.8\%$, exceeding the stronger single-component baseline, RLVR, by $2.9$ points.
The same pattern holds with external supervision: HPO surpasses Meta-Harness by $2.5$ points on SWE-Verified and $2.8$ points on average.
These consistent gains support optimizing the harness and policy together.
Appendix~\ref{app:frontier} further compares these results with frontier models on SWE-Verified.

\paragraph{The gains generalize to out-of-distribution benchmarks.}
Although harness search and policy optimization use only tasks from Python projects, the gains transfer to SWE-Multi, which covers nine non-Python languages, and SWE-Pro, which spans a broader range of languages and engineering tasks.
Self-HPO achieves $59.7\%$ and $65.2\%$, respectively, improving over Base by $5.5$ and $6.9$ points and over RLVR by $3.4$ points on both benchmarks.
The benefit is particularly clear on SWE-Multi, where Self-Harness improves over Base by only $0.1$ point, compared with $5.5$ points for Self-HPO.
The same trend holds in the external-model setup, where HPO reaches $61.4\%$ and $66.0\%$, outperforming both single-component baselines.
These results show that the benefits of joint harness--policy optimization extend beyond the Python training domain.

\paragraph{Learning from Opus 4.8 through harness optimization is more effective than SFT distillation.}
SFT and Meta-Harness represent two different approaches for transferring knowledge from Opus 4.8: SFT distills generated trajectories into the model weights, whereas Meta-Harness uses external proposals to optimize the harness while keeping the model fixed.
Meta-Harness outperforms SFT by $4.1$ points on SWE-Verified and $8.9$ points on average.
The gap is largest out of domain: SFT falls below Base on both SWE-Multi and SWE-Pro, while Meta-Harness improves on both.
HPO further improves on Meta-Harness, indicating that policy optimization remains beneficial when coupled with an optimized harness.

\begin{figure}[t]
    \centering
    \includegraphics[width=\textwidth]{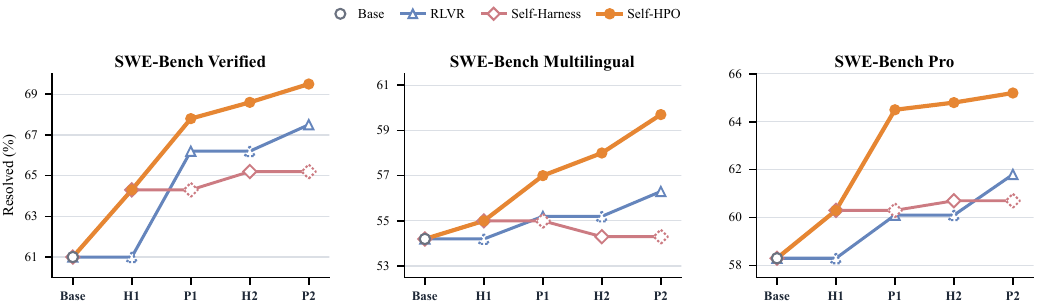}
    \caption{\textbf{Optimization trajectories across harness and policy updates.}
    Resolve rates (\%) on SWE-Bench Verified, SWE-Bench Multilingual, and SWE-Bench Pro are shown on the common Base--$H_1$--$P_1$--$H_2$--$P_2$ schedule.
    Self-HPO alternates H- and P-steps, whereas Self-Harness and RLVR update only the harness and policy, respectively.
    Markers with dashed outlines indicate steps at which the corresponding method performs no update.}
    \label{fig:ladder}
\end{figure}

\paragraph{Alternating harness and policy optimization drives continued agent improvement.}
Figure~\ref{fig:ladder} traces the optimization trajectories of Self-HPO and the two one-sided baselines.
Self-HPO and Self-Harness begin with the same first H-step, but their trajectories diverge once Self-HPO updates its policy.
Across all three benchmarks, the second H-step improves the updated Self-HPO policy, and the following P-step provides a further gain.
In contrast, another harness-only update does not consistently improve performance: Self-Harness drops from $55.0\%$ to $54.3\%$ on SWE-Multi.
The final Self-HPO pair surpasses both single-component baselines on every benchmark, supporting repeated adaptation of the harness to the evolving policy.
Figure~\ref{fig:search_frontier} in the appendix provides a candidate-level view of harness search on SWE-Verified.

\subsection{Ablations}
\label{sec:ablations}

We examine the policy-update strategy, search-set construction, and harness proposer on SWE-Verified.
Table~\ref{tab:step_ablations} uses HPO as the reference for ablating the shared H-step--P-step framework, and replacing its external proposer with the current policy gives Self-HPO.

\begin{wraptable}[8]{r}{0.44\textwidth}
\vspace{-1.5em}
\centering
\caption{\textbf{Ablation study.}
Resolve rates reported on SWE-Bench Verified.}
\label{tab:step_ablations}
\vspace{-0.6em}
\small
\setlength{\tabcolsep}{4pt}
\renewcommand{\arraystretch}{1.08}
\resizebox{\linewidth}{!}{%
\begin{tabular}{@{}lc@{}}
\toprule
\textbf{Configuration} & \textbf{Resolve rate (\%)} \\
\midrule
\textbf{HPO} & \textbf{71.6} \\
\quad w/ SFT in \pstep & 67.2\dlt{$-$4.4} \\
\quad w/ RFT in \pstep & 70.5\dlt{$-$1.1} \\
\quad w/ Fixed search set in \hstep & 70.1\dlt{$-$1.5} \\
\quad w/ Self-proposer in \hstep & 69.5\dlt{$-$2.1} \\
\bottomrule
\end{tabular}
}
\vspace{-0.5em}
\end{wraptable}
\paragraph{RL yields stronger policy updates than SFT/RFT.}
We replace RL in the P-step with either SFT on trajectories distilled from Opus 4.8 or rejection-sampling fine-tuning (RFT) on the target policy's own successful trajectories.
SFT reduces the final resolve rate from $71.6\%$ to $67.2\%$, while RFT reaches $70.5\%$, trailing RL by $1.1$ points.
Even when fine-tuning uses the policy's own successful trajectories, it still falls short of RL, supporting RL as the default \pstep.

\paragraph{Evolving the search set helps harness optimization keep pace with the policy.}
Reusing the initial search set lowers performance from $71.6\%$ to $70.1\%$.
The evolving set instead directs the next \hstep toward hard tasks from the preceding \pstep rollouts.
The gain supports refreshing the search target as the agent improves: failures relevant to an earlier policy need not remain the most useful targets for its successor.

\paragraph{Self-proposal retains strong performance without an external proposer.}
Replacing Opus 4.8 with the current target policy yields $69.5\%$, a $2.1$-point decrease from HPO.
Although external proposals provide an advantage, Self-HPO still exceeds both self-improving single-component baselines in Table~\ref{tab:main_comparison}.
Therefore, an external model is helpful but not required for our method.

\subsection{Analysis}
\label{sec:analysis}

\begin{figure}[t!]
    \centering
    \begin{subfigure}[b]{0.325\textwidth}
        \centering
        \includegraphics[width=\linewidth]{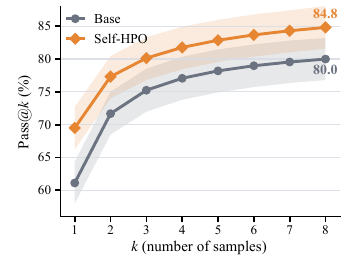}
        \caption{Pass@$k$}
        \label{fig:passk_curve}
    \end{subfigure}\hfill
    \begin{subfigure}[b]{0.325\textwidth}
        \centering
        \includegraphics[width=\linewidth]{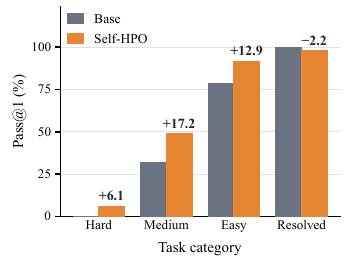}
        \caption{Pass@1 by task difficulty}
        \label{fig:pass1_breakdown}
    \end{subfigure}\hfill
    \begin{subfigure}[b]{0.325\textwidth}
        \centering
        \includegraphics[width=\linewidth]{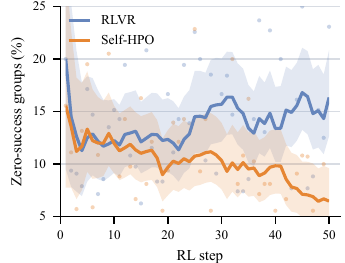}
        \caption{Zero-success groups during RL}
        \label{fig:all0_ratio}
    \end{subfigure}
    \caption{\textbf{Pass@$k$ performance on SWE-Bench Verified and RL training signal.}
    (a) pass@$k$ curve of Base and \model; (b) pass@1 by task difficulty, categorized by Base success rate $p$: Hard ($p=0$), Medium ($0<p\leq0.5$), Easy ($0.5<p<1$), and Resolved ($p=1$).
    (c) Fraction of zero-success rollout groups for RLVR and Self-HPO's final \pstep.}

    \label{fig:passk}
\end{figure}

\begin{wraptable}[10]{R}{0.44\textwidth}
\vspace{-\intextsep}
\centering
\caption{\textbf{Cross-evaluating Self-HPO policies and harnesses on SWE-Bench Verified.}
Rows vary the harness and columns vary the policy.}
\vspace{-.5em}
\small
\setlength{\tabcolsep}{5pt}
\begin{tabular}{@{}lccc@{}}
\toprule
 & \textbf{$P_0$} & \textbf{$P_1$} & \textbf{$P_2$} \\
\midrule
$H_0$ & $\tabmeanstd{61.0}{0.2}$ & $\tabmeanstd{61.7}{1.3}$ & $\tabmeanstd{64.4}{1.1}$ \\
$H_1$ & $\tabmeanstd{64.3}{1.1}$ & $\tabmeanstd{67.8}{1.6}$ & $\tabmeanstd{68.1}{1.0}$ \\
$H_2$ & $\tabmeanstd{64.5}{1.5}$ & $\tabmeanstd{68.6}{0.5}$ & $\tabbestmeanstd{69.5}{0.9}$ \\
\bottomrule
\end{tabular}
\label{tab:factorized_transfer}
\end{wraptable}

\paragraph{Both optimized components retain benefits even when paired with their base counterparts.}
Table~\ref{tab:factorized_transfer} cross-evaluates policies and harnesses.
Applying $H_2$ to the base policy improves resolve rate from $61.0\%$ to $64.5\%$, while deploying $P_2$ under the seed harness reaches $64.4\%$.
Combining $H_2$ and $P_2$ reaches $69.5\%$, an $8.5$-point gain over Base that exceeds the sum of their individual gains ($3.5+3.4=6.9$ points) by $1.6$ points.
These results show that each component retains part of the improvement on its own, while combining them gives the largest gain.

\paragraph{Self-HPO improves reliability (pass@1) and expands the capability frontier (pass@$k$).}
To examine whether the gains persist beyond pass@1, Figures~\ref{fig:passk_curve} and~\ref{fig:pass1_breakdown} compare Base and the final Self-HPO pair using eight attempts per task on SWE-Verified.
As shown in Figure~\ref{fig:passk_curve}, Self-HPO remains above Base for every $k$ from 1 to 8, and pass@8 increases from $80.0\%$ to $84.8\%$, a gain of $4.8$ points.
Figure~\ref{fig:pass1_breakdown} shows pass@1 gains of $17.2$ and $12.9$ points on Medium and Easy tasks, where Base sometimes succeeds but remains inconsistent, and a $6.1$-point gain on Hard tasks.
At the same time, Self-HPO's pass@1 decreases by $2.2$ points on the Resolved group, showing that improvement is not uniform across tasks.
Together, the gains on Medium and Easy tasks demonstrate improved reliability, while the gains on Hard tasks and the higher pass@8 show the expanded capability frontier.

\paragraph{\model reduces zero-success rollout groups during training.}
During the final Self-HPO P-step (Figure~\ref{fig:all0_ratio}), the fraction of zero-success groups, in which no rollout succeeds, declines to $6.5\%$ over the final ten updates, whereas it remains mostly between $11\%$ and $17\%$ over the last 50 updates of RLVR.
Under binary outcome rewards, such groups give GRPO no within-group reward contrast, so reducing them alleviates reward sparsity during policy optimization.

\FloatBarrier
\begin{figure}[t]
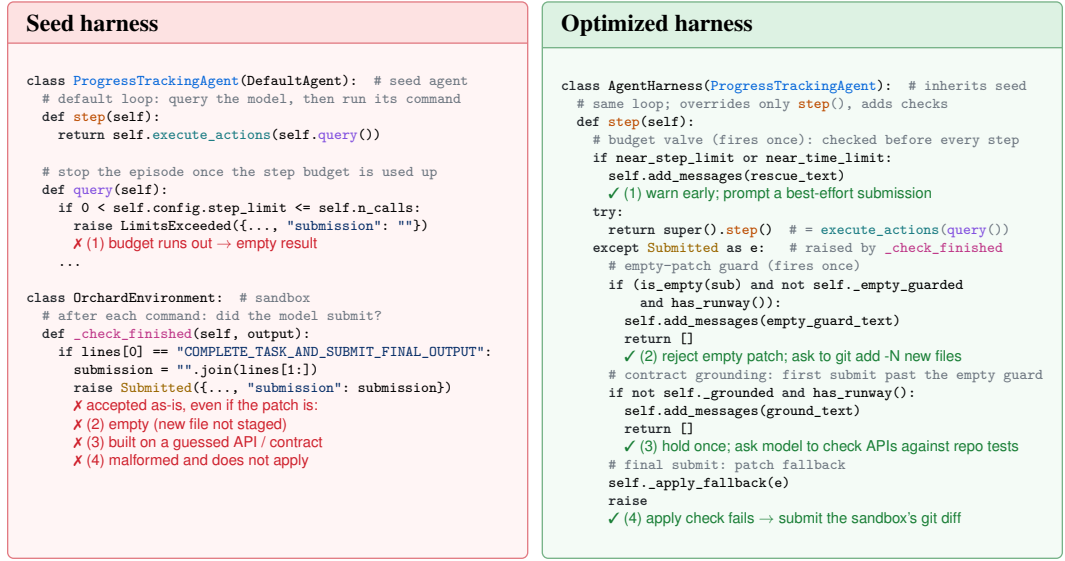

\begin{tcbraster}[raster columns=2, raster equal height=rows, raster valign=top, raster column skip=5pt]
\begin{tcblisting}{codebox, equal height group=hcode, colframe=hOld!60, colback=hOldBg, colbacktitle=hOldTitle, title={Seed harness}}
class ProgressTrackingAgent(DefaultAgent):  # seed agent
  # default loop: query the model, then run its command
  def step(self):
    return self.execute_actions(self.query())

  # stop the episode once the step budget is used up
  def query(self):
    if 0 < self.config.step_limit <= self.n_calls:
      raise LimitsExceeded({..., "submission": ""})
      |\bad{(1) budget runs out $\rightarrow$ empty result}|
    ...

class OrchardEnvironment:  # sandbox
  # after each command: did the model submit?
  def _check_finished(self, output):
    if lines[0] == "COMPLETE_TASK_AND_SUBMIT_FINAL_OUTPUT":
      submission = "".join(lines[1:])
      raise Submitted({..., "submission": submission})
      |\bad{accepted as-is, even if the patch is:}|
      |\bad{(2) empty (new file not staged)}|
      |\bad{(3) built on a guessed API / contract}|
      |\bad{(4) malformed and does not apply}|
\end{tcblisting}
\begin{tcblisting}{codebox, equal height group=hcode, colframe=hNew!60, colback=hNewBg, colbacktitle=hNewTitle, title={Optimized harness}}
class AgentHarness(ProgressTrackingAgent):  # inherits seed
  # same loop; overrides only |\cfn{cStep}{step}|(), adds checks
  def step(self):
    # budget valve (fires once): checked before every step
    if near_step_limit or near_time_limit:
      self.add_messages(rescue_text)
      |\good{(1) warn early; prompt a best-effort submission}|
    try:
      return super().step()  # = |\cfn{cExec}{execute\string_actions}|(|\cfn{cQuery}{query}|())
    except Submitted as e:   # raised by |\cfn{cCheck}{\string_check\string_finished}|
      # empty-patch guard (fires once)
      if (is_empty(sub) and not self._empty_guarded
          and has_runway()):
        self.add_messages(empty_guard_text)
        return []
        |\good{(2) reject empty patch; ask to git add -N new files}|
      # contract grounding: first submit past the empty guard
      if not self._grounded and has_runway():
        self.add_messages(ground_text)
        return []
        |\good{(3) hold once; ask model to check APIs against repo tests}|
      # final submit: patch fallback
      self._apply_fallback(e)
      raise
      |\good{(4) apply check fails $\rightarrow$ submit the sandbox's git diff}|
\end{tcblisting}
\end{tcbraster}
\caption{\textbf{Seed versus optimized harness (HPO, second H-step).}
Left: the default agent loop, with the four failure modes it lets through ({\color{hOld}\ding{55}}). Right: the optimized harness subclasses the seed agent and overrides only \texttt{step()}, and each added check ({\color{hNew}\ding{51}}) addresses the matching failure mode. Shared functions and classes use the same color on both sides. See Figure~\ref{fig:full_harness} for a detailed diff.}
\label{fig:case_layout}
\vspace{-2em}
\end{figure}

\subsection{Case Studies}
\label{sec:case_studies}

\paragraph{Harness evolution adds targeted controls around the existing agent loop.}
Figure~\ref{fig:case_layout} contrasts the seed harness with the harness obtained after HPO's second H-step.
The optimized harness retains the underlying model-query and command-execution loop, but intervenes before budget exhaustion and at submission.
First, it warns the model as the step or time limit approaches and prompts a best-effort submission before the episode ends with an empty result.
Second, a one-time empty-patch guard asks the model to include newly created files via \texttt{git add -N}.
Third, it delays the first submission that passes the empty-patch guard to prompt a check of API names and signatures against the repository and tests.
Finally, a fallback uses the sandbox's Git diff when the submitted patch fails the apply check.
These changes target distinct ways in which useful repository edits can fail to become a valid submission.
They combine explicit runtime checks, such as detecting an empty patch, with behavioral guidance, such as requesting an interface check.
Harness optimization can therefore change execution and termination logic, in addition to prompts.
Appendix~\ref{app:prompts} gives a detailed diff against the seed harness together with the injected messages, and Appendix~\ref{app:qualitative} walks through a task on which an evolved harness changes the outcome for the same policy.

\section{Related Work}
\label{sec:related}

\paragraph{Harness optimization for LLM agents.}
Harnesses govern an agent's interaction with tools and environments~\citep{yao2023react,yang2024sweagent}.
Optimization spans prompts and programs~\citep{pryzant2023protegi,khattab2023dspy,fernando2023promptbreeder,yuksekgonul2025textgrad,agrawal2026gepa}, contexts~\citep{zhang2026ace,ye2026mce,sun2026ecs}, and workflows~\citep{hu2024adas,zhang2024aflow,wang2025scoreflow,gao2025flowreasoner}.
Agents reuse feedback through refinement and reflection~\citep{madaan2023selfrefine,shinn2023reflexion}, skills~\citep{wang2023voyager}, and memory~\citep{zhao2024expel,wang2024awm}. Program evolution also improves solutions and agents~\citep{novikov2025alphaevolve,lange2025shinkaevolve,zelikman2024stop,zhang2025dgm,zhang2026hyperagents}.
Recent systems revise harnesses from execution evidence~\citep{lee2026metaharness,lin2026ahe,zhang2026selfharness,park2026autosaddler,wei2026evo,xu2026evo2}, shared experience~\citep{weng2026gea}, demonstrations~\citep{che2026demoevolve,jiang2026harnessevolve}, or self-preferences~\citep{pan2026rho}.
Other work studies continual adaptation~\citep{liu2026adaptiveautoharness,xu2026harness2}, task selection~\citep{miyai2026taskcoevolve}, generalization through search controls and modular edits~\citep{zhang2026harnesscompass,xia2026rrsi,wu2026modularrsi}, and fixed-budget evaluation~\citep{ursekar2026harnessoptbench}.
These methods adapt the agent around a fixed task-solving policy, whereas Self-HPO also updates the policy that the harness serves.

\paragraph{Policy learning and self-improvement.}
Iterative training uses filtered reasoning~\citep{zelikman2022star,singh2024restem,yuan2023scaling}, refreshed preferences~\citep{chen2024spin,yuan2024selfrewarding,pang2024irpo}, self-play curricula~\citep{zhao2025absolutezero,huang2025rzero,huang2026gzero}, negative teachers~\citep{pei2026nsd}, or evolving environments~\citep{huang2026envharness}.
Software engineering agents learn from executable environments~\citep{pan2024swegym,jain2025r2egym,yang2025swesmith}, software-evolution supervision~\citep{wei2025swerl}, and self-play code repair~\citep{wei2025selfplayswerl,choi2026anchored}.
Policy improvement also draws on verifiable-reward RL~\citep{shao2024deepseekmath,guo2025deepseek,peng2026orchard}, harness-guided distillation~\citep{ye2026harnesszero}, and test-time learning during solution search~\citep{wang2025thetaevolve,yuksekgonul2026tttdiscover}.
These methods train the policy under a fixed harness, whereas in Self-HPO the harness co-evolves with the policy being trained.

\paragraph{Joint harness--policy evolution.}
Joint adaptation spans prompts and weights~\citep{soylu2024bettertogether,tiwari2026fst,zhang2026espl,lu2026p2o}, executable harnesses and policies~\citep{hebbar2026sia,karten2026continualharness,chen2026coharness,kim2026whale,luo2026hase,chen2026harnessx,chen2026harnessforge}, and policies with evolving skills or external state~\citep{xia2026skillrl,fu2026sesa,he2026reskill,ning2026evoharness}, including safety-oriented co-evolution~\citep{mao2026safeevolve}.
MetaRSI composes data generation, harness editing, and model training within a self-improvement loop~\citep{tan2026metarsi}.
Concurrent work traces interventions~\citep{fan2026helix}, studies transfer and multi-harness training~\citep{le2026multiharness,wei2026harnessbandit}, and separates the ability to edit a harness from the ability to benefit from one~\citep{lin2026harnessbenefit}.
Self-HPO differs from these methods along two axes.
First, the policy trained in the \pstep becomes the harness proposer of the next \hstep, whereas WHALE~\citep{kim2026whale} keeps a frozen external proposer and Evo$^2$~\citep{xu2026evo2} trains a separate proposer for a frozen solver.
HASE~\citep{luo2026hase} also trains a self-proposing policy, but places harness edits in the RL action space instead of alternating harness search with RL.
Second, Self-HPO selects the next harness search set from tasks that remain unsolved during RL, whereas WHALE, Evo$^2$, and HASE search on a fixed task set.
Appendix~\ref{app:conceptual_comparison} gives a more detailed comparison with concurrent work.

\section{Conclusion}
We introduced Self-HPO, a recursive self-improving framework for coding agents that alternates harness and policy optimization.
The evolving policy proposes its own harness revisions, learns through RL under the selected harness, and supplies hard tasks for the next round of harness search.
Across three coding benchmarks, Self-HPO consistently outperforms policy-only and harness-only baselines, improving the average resolve rate from $57.8\%$ to $64.8\%$.
Analyses further show that harness and policy improvements are complementary and that Self-HPO improves reliability (pass@1) and expands the capability frontier (pass@$k$) over the base agent.

\bibliography{self-hpo}
\bibliographystyle{self-hpo}

\clearpage
\appendix
\section{Implementation Details}
\label{app:reproducibility}

\paragraph{Agent implementation.}
We use Qwen3.5-35B-A3B~\citep{qwen2026qwen35} as the backbone and SGLang~\citep{zheng2024sglang} for model serving. RL and SFT use the slime framework on one node with eight B200 GPUs.
The seed harness follows the mini-SWE-agent interface~\citep{yang2024sweagent}, through which the model executes bash commands in a task-specific repository environment.

\paragraph{Optimization data.}
\label{app:data_separation}
The initial search set $\mathcal{S}_0$ consists of 300 tasks randomly sampled from the training data $\mathcal{D}$.
For subsequent H-steps, the search set consists of hard tasks from $\mathcal{D}$, identified using the verified trajectories collected during the preceding P-step: a task is hard if it is never resolved in any of its rollouts in that P-step.
These task IDs are reused to generate fresh failure evidence with the updated policy at the next H-step.

\paragraph{Harness optimization.}
\label{app:harnesses}
Self-HPO uses the current policy checkpoint as both task solver and harness proposer.
The proposer examines the experience of evaluated harnesses and their trajectories, including failed trajectories, and revises the current best harness.
HPO instead uses Opus 4.8 to propose revisions, while retaining the same alternating H-step--P-step structure.
In both configurations, each H-step evaluates $K=6$ candidates, each run once per search task, and the selected harness is held fixed during the next policy update.
The Self-Harness and Meta-Harness baselines use the same search budget.

\paragraph{Policy optimization.}
\label{app:rl}
Each P-step initializes from the policy produced by the preceding round and collects on-policy trajectories under the updated harness.
The executable verifier supplies outcome rewards, and the policy is optimized with GRPO and Balanced Adaptive Rollout (BAR)~\citep{shao2024deepseekmath,peng2026orchard}.
Let $\mathcal{A}(\tau_i)$ denote the assistant-generated token positions in trajectory $\tau_i$, $\widehat A_i$ its group-relative advantage, and $\rho_{i,u}(\theta)$ the ratio of updated-policy to rollout-policy token probabilities.
The clipped objective is
\begin{equation}
\mathcal L_P(\theta)=-\mathbb E_i\left[
\frac{1}{|\mathcal A(\tau_i)|}\sum_{u\in\mathcal A(\tau_i)}
\min\left(\rho_{i,u}(\theta)\widehat A_i,
\operatorname{clip}(\rho_{i,u}(\theta),1-\varepsilon_{\mathrm{low}},1+\varepsilon_{\mathrm{high}})\widehat A_i\right)
\right],
\label{eq:grpo}
\end{equation}
where $\varepsilon_{\mathrm{low}}=0.2$ and $\varepsilon_{\mathrm{high}}=0.28$ are the clipping thresholds.
BAR generates eight rollouts per task and selects four to form a reward-balanced group with a success fraction between 0.25 and 0.75.
Sampling continues until 32 groups pass filtering, yielding 128 training trajectories per batch.
Each batch is optimized for one epoch with a global batch size of 128, corresponding to one optimizer update.
We use Adam with a constant learning rate of $10^{-6}$, weight decay 0.01, an entropy coefficient of $10^{-4}$, and no KL penalty.
Rollouts use temperature 0.95 and top-$p$ 0.95.
RLVR uses the same configuration under the seed harness, with 50 updates per P-step.

\section{Additional Evaluation Details}
\label{app:evaluation}

\paragraph{Evaluation configurations.}
All evaluations use a step limit of 250 and temperature and top-$p$ of 0.95.
Task attempts that fail because of sandbox or API errors are rerun, and resolve rates are computed over all tasks in each benchmark.

\paragraph{SFT configuration.}
\label{app:ablations}
The SFT baseline in Table~\ref{tab:main_comparison} fine-tunes the base policy on Opus 4.8 trajectories collected on the RL training tasks, keeping only resolved trajectories. It trains for two epochs with batch size 64 and a constant learning rate of $5\times10^{-6}$ after a 10\% warmup.
In Table~\ref{tab:step_ablations}, SFT instead replaces RL in the P-step of HPO.

\paragraph{Search-frontier curves.}
\begin{figure}[t]
    \centering
\includegraphics[width=\textwidth]{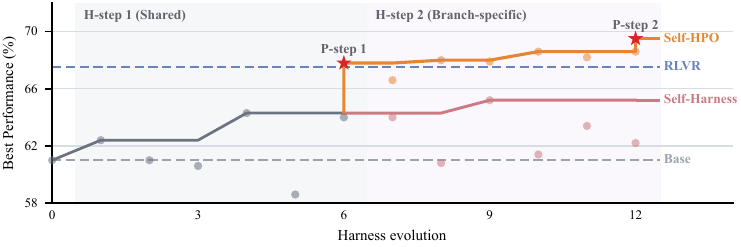}
    \caption{\textbf{Best-so-far performance during harness evolution on SWE-Bench Verified.}
    Circles show individual candidate scores.
    Candidates 1--6 are shared, while 7--12 belong to the separate Self-Harness and Self-HPO branches.
    Vertical increases and stars mark Self-HPO's two P-steps.
    Dashed horizontal lines show Base and final RLVR performance.}
    \label{fig:search_frontier}
\end{figure}
Figure~\ref{fig:search_frontier} provides a candidate-level view of harness search, complementing the selected H-step and P-step endpoints in Figure~\ref{fig:ladder}.
After the shared first H-step reaches $64.3\%$, continued harness-only search improves the best score to $65.2\%$.
Self-HPO instead updates the policy to $67.8\%$ before resuming search. Its second H-step then reaches $68.6\%$, followed by $69.5\%$ after the second policy update.
The separation between the two branches shows the benefit of policy updates alongside continued harness search.

\paragraph{Pass@$k$ calculation.}
\label{app:passk}
Figure~\ref{fig:passk} compares Base and the final Self-HPO pair with eight attempts per task.
Following~\citet{chen2021evaluating}, for a task with $c$ successes among $n=8$ attempts, we compute
\[
\operatorname{pass@}k=1-\frac{\binom{n-c}{k}}{\binom{n}{k}},\qquad k=1,\ldots,8,
\]
and average over tasks.
Pass@1 averages the success frequency, whereas pass@8 records whether any of the eight attempts succeeds.
The shaded regions show 95\% bootstrap confidence intervals.

\paragraph{Comparison with frontier models.}
\label{app:frontier}
Table~\ref{tab:frontier} compares our method with frontier models on the SWE-Bench Verified leaderboard~\citep{swebench2026leaderboard}.

\begin{table}[t]
\centering
\caption{\textbf{Comparison with frontier models on SWE-Bench Verified.} Resolve rates (\%) of frontier models on the SWE-Bench Verified leaderboard~\citep{swebench2026leaderboard} and of Qwen3.5-35B-A3B with methods from Table~\ref{tab:main_comparison}. Bold marks Self-HPO. Parameter counts are those stated in the developers' model cards (``--'' if not stated).}
\label{tab:frontier}
\small
\setlength{\tabcolsep}{2.3pt}
\begin{tabular}{llccc}
\toprule
\textbf{Model} & \textbf{Harness} & \textbf{Total Params} & \textbf{Active Params} & \textbf{Resolve Rate} \\
\midrule
\multicolumn{5}{l}{\textit{Frontier models}}\\
Gemini 3 Flash & mini-SWE-agent & -- & -- & 75.8 \\
MiniMax-M2.5 & mini-SWE-agent & -- & -- & 75.8 \\
Claude Opus 4.6 & mini-SWE-agent & -- & -- & 75.6 \\
GLM-5 & mini-SWE-agent & 744B & 40B & 72.8 \\
GPT-5.2 & mini-SWE-agent & -- & -- & 72.8 \\
Kimi K2.5 & mini-SWE-agent & 1T & 32B & 70.8 \\
DeepSeek-V3.2 & mini-SWE-agent & -- & -- & 70.0 \\
Gemini 3 Pro & mini-SWE-agent & -- & -- & 69.6 \\
Claude Haiku 4.5 & mini-SWE-agent & -- & -- & 66.6 \\
Kimi K2 Thinking & mini-SWE-agent & 1T & 32B & 63.4 \\
MiniMax-M2 & mini-SWE-agent & 230B & 10B & 61.0 \\
Devstral Small 2 & mini-SWE-agent & 24B & 24B & 56.4 \\
GPT-5 mini & mini-SWE-agent & -- & -- & 56.2 \\
Qwen3-Coder-480B-A35B & mini-SWE-agent & 480B & 35B & 55.4 \\
GLM-4.6 & mini-SWE-agent & -- & -- & 55.4 \\
Devstral 2 & mini-SWE-agent & 123B & 123B & 53.8 \\
\midrule
\multicolumn{5}{l}{\textit{Qwen3.5-35B-A3B}}\\
Base & mini-SWE-agent & 35B & 3B & 61.0 \\
SFT & mini-SWE-agent & 35B & 3B & 65.0 \\
Self-Harness & mini-SWE-agent (optimized) & 35B & 3B & 65.2 \\
RLVR & mini-SWE-agent & 35B & 3B & 67.5 \\
\textbf{Self-HPO (ours)} & \textbf{mini-SWE-agent (optimized)} & \textbf{35B} & \textbf{3B} & \textbf{69.5} \\
\bottomrule
\end{tabular}
\end{table}

\section{Conceptual Comparison with Related Methods}
\label{app:conceptual_comparison}

Table~\ref{tab:conceptual_comparison} compares Self-HPO with Meta-Harness~\citep{lee2026metaharness}, Self-Harness~\citep{zhang2026selfharness}, Task-CoEvolve~\citep{miyai2026taskcoevolve}, Evo$^2$~\citep{xu2026evo2}, WHALE~\citep{kim2026whale}, and HASE~\citep{luo2026hase} in terms of coverage of the coding domain, how the policy learns, who proposes harness changes, and how harness-search tasks are selected.

\begin{table}[!t]
\centering
\caption{\textbf{Conceptual comparison of \model and related methods.}
Coding indicates whether the method is evaluated on coding tasks.
Policy update indicates whether the task-solving policy is frozen or updated through training such as supervised fine-tuning (SFT) or reinforcement learning (RL).
Harness proposer is the model that modifies the harness, which can be either frozen or updated across rounds.
Harness search set denotes the task set used to collect diagnostic trajectories for harness optimization.
In Self-HPO, the evolving policy itself proposes harness revisions, and hard tasks identified during training form the next harness search set, keeping harness optimization targeted to the policy's evolving failure modes.}
\label{tab:conceptual_comparison}
\begingroup
\small
\setlength{\tabcolsep}{2pt}
\renewcommand{\arraystretch}{1.18}
\begin{tabular}{@{}>{\raggedright\arraybackslash}b{0.14\textwidth}>{\centering\arraybackslash}b{0.07\textwidth}>{\raggedright\arraybackslash}b{0.135\textwidth}>{\raggedright\arraybackslash}b{0.265\textwidth}>{\raggedright\arraybackslash}b{\dimexpr0.39\textwidth-20pt\relax}@{}}
\toprule
\textbf{Method} & \textbf{Coding} & \mbox{\textbf{Policy update}} & \textbf{Harness proposer} & \textbf{Harness search set} \\
\midrule
Meta-Harness & {\color{green!60!black}\cmark} & Frozen & Frozen external model & Fixed \\
\addlinespace[3pt]
Self-Harness & {\color{green!60!black}\cmark} & Frozen & Frozen self-proposer & Fixed \\
\addlinespace[3pt]
Task-CoEvolve & {\color{green!60!black}\cmark} & Frozen & Frozen external model & Evolving (high-variance tasks) \\
\addlinespace[3pt]
Evo$^2$ & {\color{green!60!black}\cmark} & Frozen & Evolving separate proposer & Fixed \\
\addlinespace[3pt]
WHALE & {\color{red!60!black}\xmark} & SFT & Frozen external model & Fixed \\
\addlinespace[3pt]
HASE & {\color{red!60!black}\xmark} & RL & Evolving self-proposer & Fixed \\
\midrule
\rowcolor{gray!15}
\textbf{Self-HPO} & {\color{green!60!black}\cmark} & \textbf{RL} & \textbf{Evolving self-proposer} & \textbf{Evolving (hard tasks from RL)} \\
\bottomrule
\end{tabular}
\endgroup
\end{table}

\paragraph{The trained solver becomes the next harness proposer.}
Self-Harness revises its harness with a fixed policy, WHALE uses a frozen external proposer alongside policy learning, and Evo$^2$ trains a separate proposer while the task-solving policy stays frozen.
Self-HPO instead reuses the updated task-solving policy as the proposer in the next round: the H-step holds policy weights fixed, and the P-step performs RL under the selected harness.
HASE also trains a self-proposing policy, but places task solving and harness editing in a joint RL action space, with persistent edits reviewed at phase boundaries.

\paragraph{Policy training also determines the next harness-search tasks.}
The P-step supplies two inputs to the next H-step: the updated proposer $P_{t+1}$ and a search set $\mathcal{S}_{t+1}$ drawn from hard tasks encountered during RL.
Thus, policy learning changes both \emph{who proposes} and \emph{which tasks guide the search}.
Task-CoEvolve also adapts task selection, but uses outcome variance across candidate harnesses within search rather than difficulty exposed during policy training.
Self-HPO updates task IDs between H-steps and keeps them fixed within each H-step.

\section{Harness Details and Prompt Templates}
\label{app:prompts}

\paragraph{Seed harness.}
The seed harness $H_0$ is the mini-SWE-agent ReAct loop~\citep{yao2023react,yang2024sweagent}: the agent issues bash tool calls in a task-specific sandbox and ends the episode by printing the patch after a fixed submission command.
Its task prompt contains the issue description followed by fixed instructions, including the exact command used to submit the patch.

\paragraph{Harness search space.}
A candidate harness is a Python class that subclasses the seed agent, together with its prompt templates.
It may override the agent's step, for example to inspect the agent's state, return feedback before accepting a submission, replace the submitted patch, or intervene at any step. All repository access goes through the sandbox.
The policy and sampling configuration (temperature and top-$p$ of 0.95) are fixed across candidates.

\paragraph{Harness proposer.}
Figure~\ref{fig:hstep_prompt} shows the system prompt of the harness proposer.
The proposer runs as a coding agent with shell access to the parent harness, its trajectories on the search set, reference fixes for search-set tasks, the outputs of earlier candidates, and notes that the proposer appends after each search round.

\par\medskip
\begin{prompt}[title={Harness Proposer System Prompt}, label=prompt:hstep, breakable]
\begin{Verbatim}[breaklines=true,breakanywhere=true,fontsize=\scriptsize]
You are the harness optimizer for a fixed policy {P_t}. The search set contains {N}
hard training tasks that the policy never solved during the preceding P-step. The
parent harness is {H_best}, the current best harness. The starting harness of this search
solves {s}% of the search set under the frozen sampling configuration. Your goal is to design ONE new harness mechanism, of your own design,
that is added on top of the parent harness and solves more search-set tasks than the
parent.

CONSTRAINTS
- Keep every method of the parent harness verbatim and add exactly one new mechanism.
- The policy, sampling configuration (temperature 0.95, top-p 0.95), step limit,
  maximum tokens, and verifier are frozen.
- Never inject reference patches or task-specific facts into the agent's context; the
  mechanism must compute its own signal at runtime.

EXPERIENCE (read it before designing; treat it as evidence, not a mandate)
- Candidates evaluated so far in this search: scores, code, and trajectories.
- Notes left by earlier proposals in this search.
Diagnostics computed from the starting harness's runs on the search set:
- Coverage gap: the source files required by the reference fix versus the files the
  agent touched.
- Precision gap: failures in which every required file was edited but the fix was
  incorrect.
- Submission loss: failures in which a real source edit appears in the trajectory but
  the final submitted patch is empty or malformed.
- Reference-fix archetypes (multi-file, signature change, new import, guard, error
  handling, ...) versus what the agent produced.
Decide for yourself which failure is the largest and most reachable by a harness, or
find another one.

WORKFLOW
0. Review the experience: the earlier notes and every candidate evaluated so far.
   For the most recent one, read the tasks it made worse and better than the starting
   harness (trajectories and reference fixes), and work out why.
1. Read several failing trajectories and compare a few with the reference fix to form
   your own hypothesis about the dominant failure cause.
2. Design a mechanism with a detectable trigger in the agent's own state and a
   concrete action: for example, intercept a submission and return feedback so that
   the agent gets another turn to act on it, replace the submitted patch, or act at
   any step.
3. Implement it in a copy of the parent harness. All repository access must go
   through the sandbox environment.
4. Write a one-line purpose (diagnosed cause -> mechanism), append what you learned
   this round to the notes, and stop. The search then checks that the candidate keeps
   every parent method and fires on real sandbox tasks, and returns any failure to you
   to fix.
\end{Verbatim}
\end{prompt}
\begingroup
\captionof{figure}{\textbf{Harness proposer system prompt} (condensed).
Values in braces change between search iterations and rounds. The parent is the current best harness, and in the first round the search set is the random $\mathcal{S}_0$ and the parent starts from the seed harness.
Implementation-specific instructions, such as file paths, the sandbox interface, and validation scripts, are omitted.}
\label{fig:hstep_prompt}
\endgroup
\medskip

\paragraph{Optimized harness.}
Figure~\ref{fig:case_layout} shows a condensed view of the harness selected in HPO's second H-step.
Figure~\ref{fig:full_harness} shows the same harness as a diff against the seed harness.
The optimized harness keeps the seed prompts unchanged and inherits the seed agent loop. Its only change is an override of \texttt{step()} that adds four checks.
Three of these checks add a message to the conversation, listed in full in Figure~\ref{fig:harness_msgs}.

\newpage
\par\medskip
\begin{tcblisting}{diffcode, title={Optimized harness (diff against the seed)}}
--- seed harness H0:     ProgressTrackingAgent.step()   (inherited agent loop)
+++ optimized harness:    AgentHarness(ProgressTrackingAgent).step()
@@ class AgentHarness(ProgressTrackingAgent) @@
 def step(self):
+    # (1) budget valve: once, shortly before the step or time limit, if nothing was submitted
+    if not self._rescued and not self._submitted_once and (near_step_limit() or near_time_limit()):
+        self._rescued = True
+        self.add_messages(user(RESCUE_TEXT))           # the step then continues as usual
-    return self.execute_actions(self.query())
+    try:
+        return self.execute_actions(self.query())      # unchanged seed step
+    except Submitted as e:                              # the policy issued the submit command
+        self._submitted_once = True
+        sub = e.submission
+        # (2) empty-patch guard: hold an empty submission once
+        if is_empty(sub) and not self._empty_guarded and has_runway():
+            self._empty_guarded = True
+            self.add_messages(user(EMPTY_GUARD_TEXT))
+            return []                                   # not submitted; the policy continues
+        # (3) contract grounding: hold the first submission that passes the empty guard once
+        if not self._grounded and has_runway():
+            self._grounded = True
+            self.add_messages(user(GROUND_TEXT))
+            return []
+        # (4) patch fallback on the final submission
+        if not is_empty(sub) and not applies_cleanly(sub):     # `git apply --check` on the base commit
+            auth = sandbox("git add -N <untracked files> && git diff")
+            if auth: e.submission = auth                # submit the sandbox's own diff instead
+        raise                                           # accept the submission

\end{tcblisting}
\begingroup
\captionof{figure}{\textbf{Harness changes from HPO's second H-step.} Shown as a diff against the seed agent's \texttt{step()} ({\color{hNew}+} added, {\color{hOld}$-$} removed). Helper calls are condensed into short names. The message texts are listed in Figure~\ref{fig:harness_msgs}.}
\label{fig:full_harness}
\endgroup
\medskip

\par\medskip
\begin{prompt}[title={(1) Budget valve: \texttt{RESCUE\_TEXT}}, breakable]
\begin{Verbatim}[breaklines=true,fontsize=\scriptsize]
[BUDGET ALMOST EXHAUSTED -- STOP EXPLORING AND SUBMIT NOW]

You are nearly out of steps/time and have not submitted a patch. An empty result scores zero, but a focused best-effort edit can pass. Right now:
1. Decide the single most likely fix location from what you already know -- do NOT start new open-ended exploration.
2. Make that concrete edit to the source file (the smallest change that matches the PR's literal request).
3. Immediately create and submit the patch, in SEPARATE commands:
   git diff -- <the source files you modified> > patch.txt
   echo COMPLETE_TASK_AND_SUBMIT_FINAL_OUTPUT && cat patch.txt
Do this in your next one or two commands.
\end{Verbatim}
\end{prompt}
\begin{prompt}[title={(2) Empty-patch guard: \texttt{EMPTY\_GUARD\_TEXT}}, breakable]
\begin{Verbatim}[breaklines=true,fontsize=\scriptsize]
[YOUR SUBMISSION IS EMPTY -- it contains no source-code changes, so it cannot fix anything. Not finalized.]

You have not edited any source file yet (patch.txt is empty or has no diff hunks). Do NOT give up. Re-localize the bug:
1. Re-read the PR and grep for the exact symbol/message/flag it names.
2. Open the file where that symbol lives and make the concrete edit the PR requires.
3. IMPORTANT -- did your fix CREATE a NEW file (a brand-new module)? A plain `git diff` does NOT show untracked new files, so your patch looks empty even though the file exists. Stage new files first so they appear as additions:
   cd /testbed && git add -N <the new file(s) you created>
4. Verify with `cd /testbed && git --no-pager diff` that a real change (or a `new file mode` hunk) now shows.
5. Then `cd /testbed && git diff -- <files> > patch.txt` and submit with:
   echo COMPLETE_TASK_AND_SUBMIT_FINAL_OUTPUT && cat patch.txt
\end{Verbatim}
\end{prompt}
\begin{prompt}[title={(3) Contract grounding: \texttt{GROUND\_TEXT}}, breakable]
\begin{Verbatim}[breaklines=true,fontsize=\scriptsize]
[GROUND YOUR DIFF IN THE REPO'S EXISTING TESTS -- one-time check; your submission is NOT finalized yet]

The most common way this task is failed is submitting a patch that APPLIES cleanly but implements the WRONG contract -- a guessed function/option name, a signature the codebase never uses, a rename or new spelling that no caller expects, or an error/return shape different from what the code around it uses. Re-reading the PR will NOT catch this, because your reading of the PR is what produced the wrong contract. The codebase itself is the authority. The repo ships EXISTING tests that already exercise the code you touched and the API next to it -- they show the EXACT names, signatures, argument spellings, and error/return types the code is expected to have. Use them as ground truth now.

1. Print your changes and identify the concrete symbols you added or changed:
   git --no-pager diff
   List every NEW or RENAMED symbol/option/message string in your diff.
2. For each such symbol, search the repo's tests and existing call sites for how that symbol -- and the nearest EXISTING sibling API -- is actually used:
   grep -rn --include='*.py' -e '<symbol>' -e '<sibling_api>' . | head -40
   Then OPEN and READ the specific test files that reference the area you changed (e.g. tests/ or test_*.py near your file) and note the exact spelling, call signature, and expected result/exception the codebase uses.
3. Confront your diff against what the tests/call sites actually expect. If you introduced a name, option spelling, signature, or error shape that NO existing test or caller uses -- while an existing convention does -- change your diff to MATCH the codebase's convention. Do not invent a new API; conform to the one already there. Also confirm you did not MISS a behavior the tests imply is required.
4. If you changed anything, regenerate the patch:
   git diff -- <the source files you modified> > patch.txt
   and re-inspect patch.txt so it contains ONLY your intended source changes.
5. When your diff conforms to the codebase's real contract, re-issue the EXACT submit command:
   echo COMPLETE_TASK_AND_SUBMIT_FINAL_OUTPUT && cat patch.txt
This check happens only once -- after you re-submit, it goes through.
\end{Verbatim}
\end{prompt}
\begingroup
\captionof{figure}{\textbf{Messages added by the optimized harness.} The messages that the harness in Figure~\ref{fig:full_harness} adds to the conversation, shown in full. Each is added at most once per task and is followed by another policy turn. The numbers match the checks in Figures~\ref{fig:case_layout} and~\ref{fig:full_harness}.}
\label{fig:harness_msgs}
\endgroup
\medskip

\section{Case Study: Where the Harness Intervenes}
\label{app:qualitative}

To illustrate how an evolved harness changes the outcome for a fixed policy, we compare two runs of the final Self-HPO policy $P_2$ on the SWE-Bench Verified task \texttt{sphinx-doc\_\_sphinx-9658}: one under the seed harness $H_0$ and one under the evolved harness $H_2$.
Here $H_2$ is the harness from Self-HPO's second H-step, which differs from the HPO harness in Appendix~\ref{app:prompts}.
We show the final steps of each run, labeling the task, the agent's reasoning, its actions, the resulting observations, and any message injected by the harness. ``$[\dots]$'' marks omitted content.
\textcolor{trajNote}{Purple notes} summarize the key steps, so the example can be followed by reading them alone.

The two runs reach the same point: the policy has edited the source, the existing tests pass, and it prepares to finish with a summary of its changes.
Under $H_0$ (Figure~\ref{fig:case_fail}), this summary becomes the final submission, so no code change is evaluated.
Under $H_2$ (Figure~\ref{fig:case_success}), the harness holds the first submission, reports that the submitted patch contains no files, and returns control to the policy, which writes the diff, checks the fix, and submits a valid patch.
The PR-intent reminder in this example is a false positive, since its keywords come from unrelated parts of the issue text.

\par\medskip\Needspace{12\baselineskip}\noindent{\raggedright\color{red!75!black}\textbf{Case 1 (Failure Case): Policy $P_2$ with the seed harness $H_0$}\par}
\begin{trajusr}[unbreakable]
\begin{Verbatim}[breaklines=true,breakanywhere=true,fontsize=\scriptsize]
<pr_description>
Consider the following PR description:
Inherited classes not correctly documented when mocked
### Describe the bug
We're experiencing an issue when documenting classes that inherit mocked classes. [...]
The base class is listed as "Bases: `torch.nn.`" instead of "Bases: `torch.nn.Module`".
[...]
</pr_description>
[task instructions of the seed harness]
\end{Verbatim}
\end{trajusr}

{\centering\scriptsize [131 earlier steps omitted]\par}
\begin{trajasst}
\begin{Verbatim}[breaklines=true,breakanywhere=true,fontsize=\scriptsize]
Good, I only modified the source files, not test files. Let me now run a final comprehensive test to make sure the fix works correctly:
\end{Verbatim}
\end{trajasst}
\begin{trajcall}
\begin{Verbatim}[breaklines=true,breakanywhere=true,fontsize=\scriptsize]
cd /testbed && python -m pytest tests/test_util_typing.py tests/test_ext_autodoc.py \
  tests/test_ext_autodoc_automodule.py tests/test_ext_inheritance_diagram.py -x --tb=short 2>&1 | tail -20
\end{Verbatim}
\end{trajcall}
\begin{trajresp}
\begin{Verbatim}[breaklines=true,breakanywhere=true,fontsize=\scriptsize]
[...]
================== 99 passed, 7 skipped, 7 warnings in 2.00s ===================
\end{Verbatim}
\end{trajresp}
\trajnote{(1) The agent's edits are in place and the existing tests pass.}
\begin{trajasst}
\begin{Verbatim}[breaklines=true,breakanywhere=true,fontsize=\scriptsize]
All tests pass. The fix is complete. Let me summarize what was done:
## Summary
[...]
\end{Verbatim}
\end{trajasst}
\begin{trajcall}
\begin{Verbatim}[breaklines=true,breakanywhere=true,fontsize=\scriptsize]
cd /testbed && echo COMPLETE_TASK_AND_SUBMIT_FINAL_OUTPUT && cat << 'EOF'
## Summary of Changes
### Issue
When documenting classes that inherit from mocked classes, the "Bases" section
incorrectly showed `torch.nn.` instead of `torch.nn.Module`.
[...]
All existing tests pass with these changes.
EOF
\end{Verbatim}
\end{trajcall}
\trajnote{(2) To finish, the agent prints a prose summary of its changes instead of the patch file.}
\begin{trajsub}
\begin{Verbatim}[breaklines=true,breakanywhere=true,fontsize=\scriptsize]
## Summary of Changes
### Issue
[...]
All existing tests pass with these changes.
\end{Verbatim}
\end{trajsub}
\trajnote{(3) The seed harness accepts this output as the submission. No code change reaches the grader, so the task is unresolved.}
\begingroup
\captionof{figure}{\textbf{Failure case: policy $P_2$ with the seed harness $H_0$.} The run ends with a prose summary as the submission, so \texttt{sphinx-doc\_\_sphinx-9658} is unresolved.}
\label{fig:case_fail}
\endgroup

\par\medskip\noindent{\raggedright\color{green!50!black}\textbf{Case 2 (Success Case): Policy $P_2$ with the evolved harness $H_2$}\par}
\begin{trajusr}
\begin{Verbatim}[breaklines=true,breakanywhere=true,fontsize=\scriptsize]
[... same task as above ...]
\end{Verbatim}
\end{trajusr}

{\centering\scriptsize [114 earlier steps omitted]\par}
\begin{trajasst}
\begin{Verbatim}[breaklines=true,breakanywhere=true,fontsize=\scriptsize]
The fix looks correct. Let me run a broader test to make sure I haven't broken anything:
\end{Verbatim}
\end{trajasst}
\begin{trajcall}
\begin{Verbatim}[breaklines=true,breakanywhere=true,fontsize=\scriptsize]
cd /testbed && python3 -m pytest tests/test_ext_autodoc*.py tests/test_util_typing.py \
  tests/test_ext_inheritance_diagram.py -x 2>&1 | tail -20
\end{Verbatim}
\end{trajcall}
\begin{trajresp}
\begin{Verbatim}[breaklines=true,breakanywhere=true,fontsize=\scriptsize]
[...]
================== 195 passed, 7 skipped, 7 warnings in 3.53s ==================
\end{Verbatim}
\end{trajresp}
\trajnote{(1) Same situation as under $H_0$: the edits are in place and the tests pass.}
\begin{trajasst}
\begin{Verbatim}[breaklines=true,breakanywhere=true,fontsize=\scriptsize]
All 195 tests pass (7 skipped). The fix is complete and working.
Let me write a summary of the fix:
## Summary
[...]
\end{Verbatim}
\end{trajasst}
\begin{trajcall}
\begin{Verbatim}[breaklines=true,breakanywhere=true,fontsize=\scriptsize]
cd /testbed && echo COMPLETE_TASK_AND_SUBMIT_FINAL_OUTPUT && cat sphinx/util/typing.py | head -130 \
  | tail -20 && echo "---" && cat sphinx/util/typing.py | tail -10
\end{Verbatim}
\end{trajcall}
\trajnote{(2) The agent again tries to finish without submitting a patch.}
\begin{trajharness}
\begin{Verbatim}[breaklines=true,breakanywhere=true,fontsize=\scriptsize]
[Full Files Auditor - Coverage Check]
----------------------------------------
You touched 1 files during exploration: ['sphinx/util/typing.py']
Your submitted patch has 0 files: (none)

FOUND GAP: 1 touched files NOT in your patch:

Potential missing files:
  - sphinx/util/typing.py

[PR Intent Reminder]
----------------------------------------
Your patch may not address the core PR intent. Key terms from PR:
  - sphinx.ext.viewcode
  - sphinx_autodoc_typehints
  - version
  - sphinxcontrib.apidoc
  - sphinx.ext.napoleon

Your submitted patch does not mention these key terms.
Consider: Does your edit actually address the issue described?
\end{Verbatim}
\end{trajharness}
\trajnote{(3) $H_2$ holds this submission once and reports that the submitted patch contains no files.}
\begin{trajasst}
\begin{Verbatim}[breaklines=true,breakanywhere=true,fontsize=\scriptsize]
The task mentioned a patch file was created. Let me verify my changes are correct and submit them properly.
\end{Verbatim}
\end{trajasst}
\begin{trajcall}
\begin{Verbatim}[breaklines=true,breakanywhere=true,fontsize=\scriptsize]
cd /testbed && git diff > patch.txt && cat patch.txt
\end{Verbatim}
\end{trajcall}
\begin{trajresp}
\begin{Verbatim}[breaklines=true,breakanywhere=true,fontsize=\scriptsize]
diff --git a/sphinx/util/typing.py b/sphinx/util/typing.py
index 87707d48f..142503422 100644
--- a/sphinx/util/typing.py
+++ b/sphinx/util/typing.py
@@ -117,6 +117,12 @@ def restify(cls: Optional[Type]) -> str:
             return cls
         elif cls in INVALID_BUILTIN_CLASSES:
             return ':py:class:`%s`' % INVALID_BUILTIN_CLASSES[cls]
+        elif _get_ismock()(cls):
+            # This is a mock object. Use its type's attributes.
+            if sys.version_info >= (3, 7):
+                return _restify_py37(type(cls))
+            else:
+                return _restify_py36(type(cls))
         elif inspect.isNewType(cls):
[...]
\end{Verbatim}
\end{trajresp}
\trajnote{(4) Prompted by the feedback, the agent writes its working-tree changes into a patch.}
\begin{trajasst}
\begin{Verbatim}[breaklines=true,breakanywhere=true,fontsize=\scriptsize]
The patch shows my changes correctly. Let me verify the fix works correctly and then submit it.
\end{Verbatim}
\end{trajasst}
\begin{trajcall}[unbreakable]
\begin{Verbatim}[breaklines=true,breakanywhere=true,fontsize=\scriptsize]
cd /testbed && python3 -c "
from sphinx.ext.autodoc.mock import _MockModule
from sphinx.util.typing import restify
mock_mod = _MockModule('torch.nn')
class DeepKernel(mock_mod.Module):
    pass
result = restify(DeepKernel.__orig_bases__[0])
print('Result:', result)
[...]"
\end{Verbatim}
\end{trajcall}
\begin{trajresp}
\begin{Verbatim}[breaklines=true,breakanywhere=true,fontsize=\scriptsize]
Result: :py:class:`torch.nn.Module`
Expected: :py:class:`torch.nn.Module`
Fix works: True
\end{Verbatim}
\end{trajresp}
\trajnote{(5) It checks the fix on the example from the issue.}
\begin{trajasst}
\begin{Verbatim}[breaklines=true,breakanywhere=true,fontsize=\scriptsize]
The fix is working correctly. Now let me submit the final output.
\end{Verbatim}
\end{trajasst}
\begin{trajcall}
\begin{Verbatim}[breaklines=true,breakanywhere=true,fontsize=\scriptsize]
echo COMPLETE_TASK_AND_SUBMIT_FINAL_OUTPUT && cat /testbed/patch.txt
\end{Verbatim}
\end{trajcall}
\enlargethispage{\baselineskip}
\begin{trajsub}
\begin{Verbatim}[breaklines=true,breakanywhere=true,fontsize=\scriptsize]
diff --git a/sphinx/util/typing.py b/sphinx/util/typing.py
[...]
\end{Verbatim}
\end{trajsub}
\trajnote{(6) The final submission is a valid patch, which resolves the task.}
\begingroup

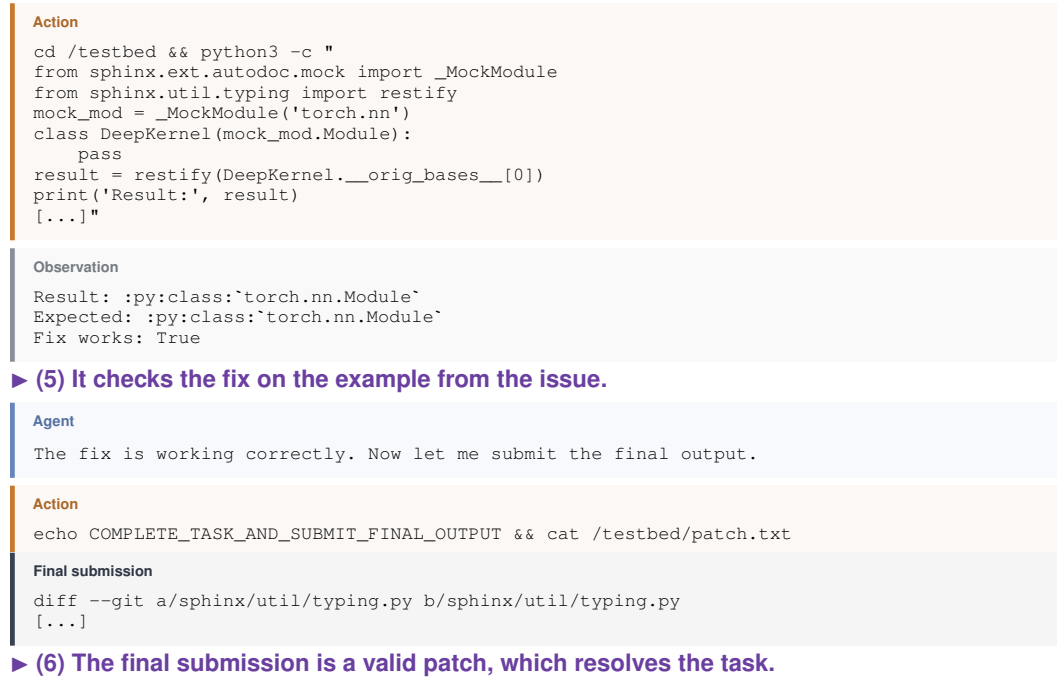
\captionof{figure}{\textbf{Success case: policy $P_2$ with the evolved harness $H_2$.} The harness holds the first submission, and the policy then submits a valid patch that resolves \texttt{sphinx-doc\_\_sphinx-9658}.}
\label{fig:case_success}
\endgroup

\end{document}